\documentclass[10pt]{article}
\usepackage[margin=0.92in]{geometry}
\usepackage[T1]{fontenc}
\usepackage[utf8]{inputenc}
\usepackage{lmodern}
\usepackage{amsmath}
\usepackage{booktabs}
\usepackage{array}
\usepackage{microtype}
\usepackage{enumitem}
\usepackage{capt-of}
\usepackage{hyperref}
\usepackage{xcolor}
\newcolumntype{L}[1]{>{\raggedright\arraybackslash}p{#1}}
\setlist{itemsep=2pt, topsep=4pt, parsep=0pt, partopsep=0pt}
\hypersetup{
    colorlinks=true,
    linkcolor=blue,
    citecolor=blue,
    urlcolor=blue,
    pdftitle={Partial Evidence Bench: Benchmarking Authorization-Limited Evidence in Agentic Systems},
    pdfauthor={Krti Tallam},
    pdfsubject={Agent evaluation, authorization-limited evidence, benchmark design},
    pdfkeywords={agentic systems, authorization, benchmarking, completeness awareness, enterprise AI}
}

\title{Partial Evidence Bench: Benchmarking Authorization-Limited Evidence in Agentic Systems}
\author{Krti Tallam\\KamiwazaAI\\\texttt{krti@kamiwaza.ai}}
\date{May 2026}

\begin{document}
\maketitle

\begin{abstract}
Enterprise agents increasingly operate inside scoped retrieval systems, delegated workflows, and policy-constrained evidence environments. In these settings, access control can be enforced correctly while the system still produces an answer that appears complete even though material evidence lies outside the caller's authorization boundary. We call this failure mode \emph{authorization-limited evidence}. It is not ordinary hallucination, not ordinary retrieval error, and not a direct permission break. It is a result-integrity failure in which the system silently overstates the completeness of what it was allowed to see. This paper introduces \texttt{partial-evidence-bench}, a deterministic benchmark for measuring that failure mode. The benchmark ships three scenario families---due diligence, compliance audit, and security incident response---with 72 tasks total, ACL-partitioned corpora, oracle complete answers, oracle authorized-view answers, oracle completeness judgments, and structured gap-report oracles. We evaluate systems along four surfaces: answer correctness, completeness awareness, gap-report quality, and unsafe completeness behavior. Checked-in baselines show that silent filtering is catastrophically unsafe across all three families, while explicit fail-and-report behavior eliminates unsafe completeness without collapsing the benchmark into a trivial abstention task. Preliminary real-model runs show model-dependent and scenario-sensitive differences in whether systems overclaim completeness, conservatively underclaim, or express incompleteness in an enterprise-usable form. The broader contribution is to make a governance-critical agent failure measurable without human judges or contamination-prone static corpora.
\end{abstract}

\section{Introduction}

Agentic AI systems are moving from single-prompt assistants toward retrieval-backed, tool-using, delegated workflows that act inside real organizational boundaries. Contemporary systems increasingly combine retrieval-augmented generation \cite{lewis2020rag}, tool use \cite{schick2023toolformer}, and action loops over external environments \cite{yao2023react}. That shift is exactly why a narrower evaluation question now matters: what happens when the system's view of the world is intentionally partial, but the task still invites a decisive answer?

Enterprise environments make this concrete. The acting system may see a deal room but not privileged annexes, an audit workspace but not exception-management compartments, or an incident-response bridge but not restricted forensic or legal material. In all of these cases, the access-control layer can be functioning exactly as designed. The harder question is what the system does next, and more specifically what it implicitly communicates about the completeness of its result.

The usual benchmark framing is not sufficient here. Most evaluation work asks whether a system can retrieve the right document, answer the task correctly, or avoid violating explicit permissions. Those questions matter, but they miss a narrower and operationally important failure mode: the visible evidence can be coherent enough to support a polished synthesis even when it is materially incomplete. The result is not an unauthorized access violation. It is an answer that looks complete to the consumer while silently reflecting only the authorized subset of the truth.

This paper argues that \emph{authorization-limited evidence} is a benchmarkable failure surface. The central issue is not whether a model can break permissions, but whether it can recognize when its answer is incomplete because relevant evidence lies outside the caller's authorization boundary, and whether it can report that incompleteness explicitly enough to be operationally useful. That matters because enterprise deployment decisions are often not made on raw model output quality alone. They are made on whether a system can preserve result integrity under policy constraints.

To make that claim precise, we introduce \texttt{partial-evidence-bench}, a deterministic benchmark built around synthetic corpora with known ground truth, explicit ACL partitions, and scenario-specific completeness oracles. The benchmark evaluates systems along four separate surfaces:
\begin{enumerate}
    \item answer correctness,
    \item completeness awareness,
    \item gap-report quality, and
    \item unsafe completeness behavior.
\end{enumerate}

The design is influenced by deterministic evaluation methodology such as RIKER's ground-truth-first construction \cite{valencia2026riker}, but targets a different question: not general retrieval quality, but correctness under intentionally authorization-limited evidence views. It also complements broad agent benchmarks such as AgentBench \cite{liu2023agentbench} and KAMI-style agentic evaluations \cite{valencia2026kami} by focusing on a result-integrity failure that can arise even when the system never attempts to cross a permission boundary.

This paper makes five contributions:
\begin{enumerate}
    \item It defines authorization-limited evidence as a benchmark problem distinct from both ordinary hallucination and direct access-control violation.
    \item It introduces a deterministic benchmark with three built-in scenario families and 72 total tasks.
    \item It proposes a metric suite that separates local answer quality from completeness-awareness behavior.
    \item It shows, through checked-in baselines, that silent filtering is unsafe across all shipped families.
    \item It reports preliminary real-model findings showing that the failure mode is model-dependent and scenario-sensitive.
\end{enumerate}

\section{Why This Failure Mode Matters Now}

Three current shifts in agentic AI make this benchmark timely rather than niche.

\paragraph{First, the action surface has widened.}
Modern agent systems do not merely answer questions from static prompts. They traverse corpora, call tools, produce summaries for downstream humans, and increasingly operate inside multi-step workflows. That makes it more dangerous to collapse evaluation into raw answer accuracy. A locally plausible answer can still be operationally unsafe if it misrepresents the limits of the evidence that was available.

\paragraph{Second, enterprise deployments are boundary-rich by design.}
Real deployments are full of intentional scope cuts: role-based access control, matter partitions, legal holds, privileged channels, country restrictions, and incident compartments. A system can therefore be fully compliant at the access layer while still producing a result that encourages an over-broad downstream action. Related governance work on fail-and-report \cite{tallam2026failreport}, authorization propagation \cite{tallam2026authprop}, and execution-envelope admission \cite{tallam2026execenv} argues that identity, boundary, and admissibility must remain explicit as requests flow through agentic systems. \texttt{partial-evidence-bench} provides an evaluation surface for one specific downstream consequence of getting that wrong.

\paragraph{Third, current evaluation literature leaves a gap.}
Hallucination and behavioral-testing work such as HELM \cite{liang2023helm}, CheckList \cite{ribeiro2020checklist}, and unanswerability settings such as SQuAD~2.0 \cite{rajpurkar2018know} ask important questions, but they do not directly test whether a system can represent the incompleteness induced by authorization boundaries. A task can be answerable in the full world, partially answerable in the authorized world, and still dangerous if the system presents the partial world as the whole world. That is the narrower issue this benchmark isolates.

The paper's position is therefore straightforward: authorization-limited evidence is not a corner case of general uncertainty. It is a first-class agent evaluation problem for systems that act over scoped enterprise evidence.

\section{Problem Definition}

Let $D$ denote the full document set relevant to a task, and let $A \subseteq D$ denote the authorized view available to the acting system. Let $q$ be the task prompt and let $\mathcal{R}(A, q)$ denote the system response. In many enterprise settings, $A$ is not an arbitrary sample; it is a policy-constrained projection of $D$ created by authorization boundaries.

The benchmark focuses on tasks where:
\begin{enumerate}
    \item the full-corpus answer under $D$ is known,
    \item the authorized-view answer under $A$ is known,
    \item the required evidence families for a complete answer are known, and
    \item the correct completeness judgment under $A$ is known.
\end{enumerate}

This yields three important distinctions.

\paragraph{Correctness under authorization.}
If the task remains complete under $A$, the correct answer is the same as the full-corpus answer. If the task becomes incomplete under $A$, the benchmark may still permit a partial answer, but that answer is judged against the authorized-view oracle rather than the full-corpus oracle.

\paragraph{Completeness awareness.}
The system must correctly represent whether the visible evidence base is complete enough for the task being asked. This is not reducible to answer quality alone.

\paragraph{Unsafe completeness behavior.}
The headline failure occurs when the system returns a complete-seeming answer for a materially incomplete task without emitting a correct gap report.

Formally, over the subset $\mathcal{I}$ of materially incomplete tasks, the benchmark defines:
\[
\mathrm{unsafe\_completeness\_rate}
=
\frac{
\left|
\left\{
t \in \mathcal{I} \,:\,
\mathcal{R}(A_t, q_t)\ \text{appears complete and lacks a correct gap report}
\right\}
\right|
}{|\mathcal{I}|}.
\]

This is the benchmark's primary metric because it directly captures the operational harm: the consumer is misled about completeness, not merely given a low-quality answer.

\section{Benchmark Construction}

\subsection{Scenario Families}

The benchmark currently ships three scenario families, each with six task templates and four fixed seeds, for a total of 72 tasks. Each family is designed to preserve deterministic scoring while applying different pressure toward false completeness.

\begin{table}[t]
\centering
\small
\begin{tabular}{L{0.22\linewidth} L{0.27\linewidth} L{0.24\linewidth} L{0.17\linewidth}}
\toprule
\textbf{Scenario} & \textbf{Decision setting} & \textbf{Hidden evidence pattern} & \textbf{Tasks} \\
\midrule
\texttt{due\_diligence\_v1} & Transaction diligence and recommendation & Hidden annexes, contradictory analysis, forecast downgrades, concentration signals & 24 \\
\texttt{compliance\_audit\_v1} & Audit readiness and control assessment & Exception registers, override logs, retention failures, access drift & 24 \\
\texttt{security\_incident\_v1} & Incident containment and disclosure posture & Scope expansion, regulated data exposure, credential abuse, legal hold triggers & 24 \\
\bottomrule
\end{tabular}
\caption{Shipped scenario families in \texttt{partial-evidence-bench}.}
\label{tab:families}
\end{table}

The third family strengthens the benchmark's coverage because it includes contradiction-bearing hidden evidence rather than only missing annexes or latent exception records. This matters for real-world agent behavior: some failures arise from lack of corroboration, while others arise from a visible narrative being actively falsified by restricted material.

\begin{table}[t]
\centering
\small
\begin{tabular}{L{0.22\linewidth} c r r r r r}
\toprule
\textbf{Scenario} & \textbf{Patterns} & \textbf{Complete} & \textbf{Incomplete} & \textbf{Partial} & \textbf{Block} & \textbf{Hidden-outside} \\
\midrule
\texttt{due\_diligence\_v1} & $6 \times 4$ & 3 & 21 & 12 & 12 & 21 \\
\texttt{compliance\_audit\_v1} & $6 \times 4$ & 4 & 20 & 16 & 8 & 20 \\
\texttt{security\_incident\_v1} & $6 \times 4$ & 7 & 17 & 8 & 16 & 17 \\
\bottomrule
\end{tabular}
\caption{Task composition by shipped scenario family. ``Patterns'' denotes six hidden-evidence pattern families instantiated at four fixed seeds. ``Partial'' and ``Block'' denote whether \texttt{partial\_response\_permitted} is true or false for the task contract. ``Hidden-outside'' counts tasks where material evidence remains outside the authorized view.}
\label{tab:split}
\end{table}

The split matters because the benchmark is not a pure abstention test. Each family contains both cases where the hidden material is promoted into the authorized view and cases where it remains outside scope. It also mixes incomplete tasks where structured partial synthesis is acceptable with incomplete tasks where policy requires a blocked response. The benchmark is therefore testing scoped reasoning behavior, not merely whether a model learned to refuse whenever uncertainty appears.

\subsection{Design Criteria}

The benchmark was designed against four constraints that matter in current agent evaluation.

\paragraph{Deterministic scoring.}
The benchmark should not require human graders to decide whether an answer was ``complete enough.'' Instead, completeness should be derivable from known document structure, required evidence families, and scenario-specific oracles.

\paragraph{Policy realism without live enterprise data.}
The benchmark should capture realistic access-boundary patterns without depending on private corpora that cannot be redistributed or whose ground truth is inherently ambiguous. Synthetic corpora are therefore not a convenience here; they are what make the target failure measurable at all.

\paragraph{Non-trivial partial-answer conditions.}
A useful benchmark cannot collapse into universal abstention. Some tasks must remain fully answerable under the authorized view, while others must become incomplete in ways that still permit structured partial reporting. The benchmark therefore mixes complete and incomplete cases within each family.

\paragraph{Enterprise-usable incompleteness signals.}
The benchmark should reward more than a vague hedge. In realistic deployments, downstream humans need a reusable explanation of what boundary or missing evidence made the answer incomplete. That is why gap-report quality is scored as its own surface rather than being treated as a stylistic bonus.

\subsection{Document and ACL Construction}

Each task instance contains six synthetic documents:
\begin{enumerate}
    \item an overview memo,
    \item a primary evidence packet,
    \item a public or policy context note,
    \item an advisor or coordinator note,
    \item a hidden material document, and
    \item a neutral tracker document.
\end{enumerate}

The benchmark assigns scopes to documents and defines a user scope for the acting system. Most tasks deliberately place the hidden material outside the authorized scope, but a deterministic subset of seed-template pairs promotes the hidden material into the authorized scope. This creates both complete and incomplete tasks within the same family and avoids collapsing the benchmark into pure abstention detection.

For each task, the generator emits:
\begin{itemize}
    \item the full document set,
    \item the authorized document ids,
    \item required evidence families,
    \item a complete-answer oracle,
    \item an authorized-view answer oracle, and
    \item an expected gap report oracle.
\end{itemize}

\subsection{Worked Example}

Table~\ref{tab:workedexample} shows one representative due-diligence task, \texttt{hidden\_liability\_annex\_\_s11}. It is useful because the visible evidence is coherent and points toward a cautious proceed recommendation, while the hidden document flips the correct complete-world answer to a block. This is exactly the kind of case where a polished scoped summary is dangerous if the system does not surface the missing boundary.

\begin{center}
\small
\begin{tabular}{L{0.24\linewidth} L{0.68\linewidth}}
\toprule
\textbf{Field} & \textbf{Value} \\
\midrule
Prompt & Review the liability posture around Westbridge MedTech exposure and state whether the transaction should proceed. \\
\midrule
Authorized visible evidence & Advisor memo: transaction appears supportable subject to contractual complexity. Data-room update: visible posture appears stable. Public note: no material issue is visible from public disclosures. Tracker: visible workstreams are complete, but restricted internal compartments are excluded. \\
\midrule
Hidden evidence outside scope & Internal liability annex memo in scope \texttt{target\_internal} showing \texttt{uninsured\_liability\_annex}, which materially changes the diligence posture. \\
\midrule
Oracle complete answer & Recommendation: \texttt{block}; material flag: \texttt{liability\_annex\_missing}; risk summary: \texttt{contractual\_complexity}, \texttt{uninsured\_liability\_annex}. \\
\midrule
Oracle authorized-view answer & Recommendation: \texttt{proceed\_with\_caution}; no material flags; risk summary: \texttt{contractual\_complexity}. \\
\midrule
Expected gap report & Status: \texttt{blocked}; missing requirement: \texttt{liability\_annex}; missing scope: \texttt{target\_internal}; affected claims: recommendation, risk summary, and material flags; reason: \texttt{policy\_requires\_block}. \\
\midrule
\texttt{silent\_filter} behavior & Returns the authorized-view answer, emits no gap report, and therefore creates an unsafe completeness failure: a decisive recommendation is given even though the required liability evidence is outside scope. \\
\bottomrule
\end{tabular}
\captionof{table}{Worked example from \texttt{due\_diligence\_v1}: task \texttt{hidden\_liability\_annex\_\_s11}.}
\label{tab:workedexample}
\end{center}

Two aspects of this example are important. First, the visible evidence is not absurd or obviously contradictory; a model that answers from only the authorized view can still sound reasonable. Second, the tracker document makes the hiddenness legible in principle by stating that restricted compartments are excluded. This means the benchmark is not asking the model to infer magical unseen documents from nothing. It is asking whether the model can treat an explicit boundary signal as a first-class reason to qualify or block its answer.

\subsection{Why Deterministic Synthetic Corpora}

Static enterprise datasets make it difficult to know whether a model is failing because of contamination, retrieval variance, or scoring ambiguity. Following the ground-truth-first logic used in RIKER-style evaluation \cite{valencia2026riker}, this benchmark generates documents from known semantic structure rather than trying to infer ground truth after the fact. That enables:
\begin{itemize}
    \item deterministic correctness scoring,
    \item deterministic completeness judgments,
    \item reproducible ACL partitions, and
    \item contamination-resistant scenario regeneration.
\end{itemize}

This choice also matters scientifically. In many enterprise-like settings, ``the right answer'' is only obvious after a reviewer reconstructs which documents should have been considered and which ones were out of scope. That makes the benchmark target easy to argue about but hard to measure. By emitting both a full-world oracle and an authorized-world oracle, \texttt{partial-evidence-bench} turns that ambiguity into an explicit, testable distinction.

\section{Scoring}

The benchmark keeps four surfaces separate.

\subsection{Answer Correctness}

Answer correctness is field-based and deterministic. It does not always compare the model output against the full-corpus oracle. Instead:
\begin{itemize}
    \item if the task is complete under authorization, correctness is scored against the complete answer;
    \item if the task is incomplete but partial response is allowed, correctness is scored against the authorized-view answer;
    \item if the task is incomplete and policy requires blocking, the correct answer is no answer.
\end{itemize}

\subsection{Completeness Awareness}

Completeness awareness is a categorical judgment over whether the system correctly recognizes task completeness under the authorized evidence view. The benchmark tracks:
\begin{itemize}
    \item \texttt{correct\_complete},
    \item \texttt{correct\_incomplete},
    \item \texttt{false\_complete}, and
    \item \texttt{false\_incomplete}.
\end{itemize}

Separating this from answer correctness is essential. A system may be locally correct relative to the documents it can see and still fail to recognize that those documents are insufficient for the actual task. Conversely, a system may be overly conservative and declare incompleteness when the authorized view was already sufficient. Those are different deployment risks and they should not be collapsed into one scalar.

\subsection{Gap-Report Quality}

Gap-report quality measures how well the system describes what is missing and why. It is computed using deterministic overlap across:
\begin{itemize}
    \item missing scopes,
    \item missing requirements, and
    \item affected claims.
\end{itemize}

This matters because enterprise consumers often need more than a vague refusal or caveat. They need a reusable, structured statement of what evidence boundary made the answer incomplete.

\subsection{Unsafe Completeness Behavior}

Unsafe completeness is the benchmark's headline metric because it captures the central failure directly. A system can have locally plausible correctness, perfect citation honesty on visible documents, and still fail catastrophically if it presents an incomplete result as complete.

\subsection{Why These Four Surfaces Belong Together}

Taken together, the metric suite distinguishes several practically important operating profiles:
\begin{itemize}
    \item systems that answer well and know when they are incomplete,
    \item systems that answer plausibly but overclaim completeness,
    \item systems that avoid unsafe completeness by blocking too often, and
    \item systems that identify incompleteness but fail to explain it in a reusable way.
\end{itemize}

That decomposition is one of the benchmark's main design claims. In scoped enterprise settings, answer quality without completeness awareness is not enough, and completeness awareness without an intelligible gap report is often not enough either.

\section{Baseline Behaviors}

The benchmark ships four built-in baseline adapters:
\begin{enumerate}
    \item \texttt{silent\_filter},
    \item \texttt{warning\_partial},
    \item \texttt{fail\_and\_report}, and
    \item \texttt{oracle}.
\end{enumerate}

\texttt{silent\_filter} always answers from the visible evidence and never treats missing evidence as a first-class condition. \texttt{warning\_partial} always answers but emits an incomplete gap report when the task is incomplete. \texttt{fail\_and\_report} blocks when policy requires blocking and otherwise emits a fully structured gap report. \texttt{oracle} provides a deterministic upper bound and is expected to match \texttt{fail\_and\_report} on the shipped tasks.

\begin{table}[t]
\centering
\small
\begin{tabular}{L{0.23\linewidth} L{0.18\linewidth} r r r}
\toprule
\textbf{Scenario} & \textbf{Baseline} & \textbf{Unsafe} & \textbf{Answer} & \textbf{Gap} \\
\midrule
due diligence & silent filter & 1.000 & 0.583 & 0.125 \\
due diligence & warning partial & 0.000 & 0.583 & 0.833 \\
due diligence & fail-and-report & 0.000 & 1.000 & 1.000 \\
\midrule
compliance audit & silent filter & 1.000 & 0.667 & 0.167 \\
compliance audit & warning partial & 0.000 & 0.667 & 0.867 \\
compliance audit & fail-and-report & 0.000 & 1.000 & 1.000 \\
\midrule
security incident & silent filter & 1.000 & 0.583 & 0.292 \\
security incident & warning partial & 0.000 & 0.583 & 0.833 \\
security incident & fail-and-report & 0.000 & 1.000 & 1.000 \\
\bottomrule
\end{tabular}
\caption{Checked-in baseline results. ``Unsafe'' denotes \texttt{unsafe\_completeness\_rate}; ``Answer'' and ``Gap'' denote average answer correctness and gap-report quality.}
\label{tab:baselines}
\end{table}

Two observations matter immediately. First, silent filtering is catastrophically unsafe across all three families. Second, the benchmark does not reward trivial refusal. The middle baseline avoids unsafe completeness but does not match the fully structured intervention on either answer quality or gap-report quality.

These baselines are deliberately simple, but they perform an important methodological role. They show that the benchmark is not merely rediscovering answer accuracy under a different name. If it were, the silent-filter baseline would remain competitive as long as the visible evidence supported polished local synthesis. Instead, it fails on exactly the surface the benchmark is meant to expose.

\section{Preliminary Real-Model Findings}

\subsection{Exploratory Multi-Model Results}

The repository includes an exploratory real-model result set over the first two scenario families using five models. The checked-in set covers three Claude variants together with \texttt{gpt-4o} and \texttt{meta-llama-3.1-8b-instruct}. These results are not yet fully transport-normalized across providers, so they should be read as early evidence rather than a final leaderboard.

\begin{table}[t]
\centering
\small
\begin{tabular}{L{0.24\linewidth} L{0.20\linewidth} r r r}
\toprule
\textbf{Model} & \textbf{Scenario} & \textbf{Unsafe} & \textbf{Answer} & \textbf{Gap} \\
\midrule
\texttt{claude\_opus} & due diligence & 0.000 & 0.677 & 0.740 \\
\texttt{claude\_sonnet} & due diligence & 0.000 & 0.586 & 0.533 \\
\texttt{gpt-4o} & due diligence & 0.048 & 0.469 & 0.650 \\
\texttt{meta-llama-3.1-8b} & due diligence & 0.000 & 0.417 & 0.350 \\
\texttt{claude\_haiku} & due diligence & 0.286 & 0.488 & 0.376 \\
\midrule
\texttt{claude\_opus} & compliance audit & 0.000 & 0.617 & 0.767 \\
\texttt{claude\_haiku} & compliance audit & 0.000 & 0.584 & 0.567 \\
\texttt{claude\_sonnet} & compliance audit & 0.000 & 0.565 & 0.533 \\
\texttt{gpt-4o} & compliance audit & 0.000 & 0.333 & 0.592 \\
\texttt{meta-llama-3.1-8b} & compliance audit & 0.000 & 0.333 & 0.379 \\
\bottomrule
\end{tabular}
\caption{Exploratory real-model results from the checked-in mixed-transport runs.}
\label{tab:realmodels}
\end{table}

These runs support three claims.

\paragraph{Silent incompleteness is separable from local answer quality.}
\texttt{claude\_haiku} still produces non-trivial answer correctness on due diligence tasks, but also incurs a high unsafe completeness rate. That is exactly the failure mode the benchmark is meant to surface.

\paragraph{Structured incompleteness reporting is a distinct capability.}
\texttt{gpt-4o} and \texttt{claude\_sonnet} are a useful contrast. Their answer-quality and gap-quality profiles differ, showing that local task competence and enterprise-usable completeness reporting do not move together automatically.

\paragraph{Scenario pressure matters.}
\texttt{claude\_haiku} is safe on compliance audit but unsafe on due diligence. That suggests the benchmark is measuring more than a single global caution scalar; different scenario families create different pressure toward false completeness.

\subsection{Interpretation}

The real-model results show that the benchmark is not simply dividing systems into safe and unsafe. It distinguishes at least three operating profiles:
\begin{enumerate}
    \item unsafe polished overclaiming,
    \item conservative underclaiming, and
    \item high-quality structured completeness awareness.
\end{enumerate}

That distinction matters for deployment. A model that never overclaims but routinely under-answers may be acceptable in some settings and unusable in others. A model that answers fluently while occasionally presenting an incomplete result as complete is often worse, because the failure is harder for the consumer to detect.

\subsection{Provider-Consistent Claude Extension}

To tighten the real-model story, we additionally evaluated all three Claude variants on the \texttt{security\_incident\_v1} family through the same \texttt{claude -p} transport path. This removes the main transport caveat from the cross-provider table above and gives a cleaner within-family comparison.

\begin{table}[t]
\centering
\small
\begin{tabular}{L{0.24\linewidth} r r r r}
\toprule
\textbf{Model} & \textbf{Unsafe} & \textbf{Answer} & \textbf{Gap} & \textbf{Blocked-safe} \\
\midrule
\texttt{claude\_opus} & 0.000 & 0.698 & 0.792 & 0.000 \\
\texttt{claude\_sonnet} & 0.000 & 0.673 & 0.617 & 0.000 \\
\texttt{claude\_haiku} & 0.000 & 0.595 & 0.563 & 0.214 \\
\bottomrule
\end{tabular}
\caption{Provider-consistent Claude-only results on \texttt{security\_incident\_v1}. ``Blocked-safe'' denotes the rate of blocking when partial answering would have been permitted.}
\label{tab:claudeincident}
\end{table}

This tighter sweep sharpens the story in two ways.

\paragraph{The third family is not redundant.}
The incident family preserves the same broad ordering as the earlier scenarios---Opus strongest, Sonnet in the middle, Haiku weakest---but it changes the failure shape. Haiku is not unsafe here in the false-complete sense. Instead, it over-blocks, with \texttt{blocked\_when\_safe\_to\_answer\_rate = 0.214}. That is useful because it shows the benchmark can distinguish between unsafe overclaiming and conservative over-suppression.

\paragraph{Structured completeness awareness remains separable from answer quality.}
All three Claude models avoid false completeness on the incident family, but their answer quality and gap-report quality remain materially different. Opus remains strongest on both, while Sonnet is meaningfully closer to Haiku than to Opus on gap quality. That supports the claim that enterprise-usable incompleteness reporting is its own capability surface rather than a free byproduct of general competence.

\section{Discussion}

\subsection{What This Benchmark Adds to Current Agent Evaluation}

The benchmark contributes a failure surface that becomes visible only when three ingredients are present at once:
\begin{enumerate}
    \item retrieval or document-grounded reasoning,
    \item authorization-constrained evidence views, and
    \item downstream consumers who may act on the apparent completeness of an answer.
\end{enumerate}

This combination is increasingly common in enterprise agent design. Yet most benchmark suites still emphasize either broad capability, generic factuality, or explicit safety-policy violations. \texttt{partial-evidence-bench} instead targets a narrower but operationally important result-integrity question: did the system preserve the distinction between what is true in the full world and what was justified in the authorized world?

That distinction matters because scoped agent systems are now being inserted into review, triage, and recommendation loops where the next actor is a human who may reasonably infer more completeness than the system actually had. In other words, the benchmark does not merely ask whether the model answered well. It asks whether the model preserved the informational preconditions under which its answer should be trusted. That is a stronger and more deployment-relevant demand.

The benchmark therefore fills a gap between classical NLP evaluation and systems governance. Classical evaluation often asks whether a model predicted the right answer. Governance work often asks whether the model crossed a boundary it should not have crossed. But enterprise failures are often subtler: the model respects the boundary, yet produces an output whose tone, structure, or decisiveness encourages the consumer to forget that the boundary existed. \texttt{partial-evidence-bench} turns that intermediate failure mode into something measurable.

\subsection{Relation to Retrieval and Hallucination Evaluation}

This benchmark is adjacent to, but not reducible to, ordinary retrieval or hallucination evaluation. It is possible for a system to:
\begin{itemize}
    \item retrieve perfectly within the authorized set,
    \item cite only visible documents honestly, and
    \item still mislead the consumer by silently omitting material evidence that exists outside the authorized boundary.
\end{itemize}

That is why authorization-limited evidence should not be collapsed into generic answer quality or citation faithfulness. It is closer to a boundary-conditioned integrity failure than to a simple factuality failure.

This point is easy to miss in current agent discourse because retrieval performance is often used as a proxy for epistemic adequacy. If the system surfaced the most relevant accessible documents and reasoned coherently over them, the output is often treated as ``well-grounded.'' But grounded relative to what? In a scoped environment, the relevant issue is not only whether the answer was grounded in the visible evidence, but whether the answer correctly represented the limits of that evidence. A system can be well grounded and still be misleading.

This is also why citation-heavy interfaces are not a complete fix. Citations can prove that the system did not fabricate its visible basis. They do not by themselves prove that the visible basis was sufficient for the task. The benchmark's emphasis on completeness awareness therefore extends rather than replaces ordinary retrieval and hallucination evaluation.

\subsection{Relation to Unanswerability and Abstention}

The benchmark also differs from classic unanswerable-question settings such as SQuAD 2.0 \cite{rajpurkar2018know}. Here, the task may be answerable in the full world and partially answerable in the authorized world. The issue is not merely whether the system abstains. It is whether it knows when its visible world is insufficient and whether it represents that insufficiency in a structured way.

This puts the benchmark in contact with recent work on self-knowledge and abstention in language models \cite{kadavath2022mostly,madhusudhan2024notanswer}, but under a different regime. Those papers ask whether models know when they do not know in a more general sense. \texttt{partial-evidence-bench} asks a sharper question: can a model represent incompleteness correctly when the missingness is induced by authorization boundaries rather than by general lack of knowledge?

That difference matters because enterprise decisions rarely reduce to a binary answerable/unanswerable split. In practice, the acting system may be able to provide a locally useful partial synthesis, but only if it also makes the missing evidence explicit enough for a downstream reviewer to interpret the result correctly. A benchmark that rewards only abstention would therefore miss an important operational distinction between safe partiality and unsafe false completeness.

Conversely, a benchmark that rewards only answer quality would push systems toward polished overclaiming. The space of desirable behavior is therefore structured: sometimes the right action is to answer with scoped caveats, sometimes it is to escalate, and sometimes it is to refuse. One value of \texttt{partial-evidence-bench} is that it begins to separate these behaviors rather than treating them all as generic uncertainty handling.

\subsection{From Permission Safety to Result Integrity}

A recurring mistake in enterprise AI discussions is to treat access control as the main governance line. Access control is necessary, but it is not sufficient. A system can satisfy every permission check in the pipeline and still fail at the moment of communication by presenting a scoped view as though it were the whole relevant world.

The benchmark therefore suggests a shift in framing. Instead of thinking only in terms of ``did the agent access something forbidden?,'' we also need to ask ``did the agent preserve the truth conditions under which its answer should be interpreted?'' This is what makes authorization-limited evidence a result-integrity problem rather than merely a security-control problem.

That framing is important for agentic AI because the surrounding conversation is often polarized between capability and safety. The failure mode here sits between them. It is a capability failure in one sense, because the system failed to reason correctly about the limits of its evidence. But it is also a governance failure, because the surrounding system failed to require a representation of those limits that downstream humans could act on safely.

\subsection{Implications for Agent and Platform Design}

The benchmark suggests a concrete systems lesson: correct authorization alone is not enough. Systems that operate over scoped corpora need a visible completeness layer. In practice, that means:
\begin{itemize}
    \item explicit gap reports,
    \item missing-scope signaling,
    \item policy-aware blocking when partial answers are unsafe, and
    \item UX surfaces that make evidence boundaries legible to the consumer.
\end{itemize}

This is especially important for scoped agents, audit environments, diligence workflows, and incident-response settings where the user may reasonably assume the system is seeing ``everything that matters'' when it is not. More broadly, the benchmark suggests that authorization and evaluation should not remain separate concerns: governance primitives such as fail-and-report \cite{tallam2026failreport} need matching evaluation surfaces if they are going to shape real deployment choices.

The design implication is not just ``add a disclaimer.'' In many settings, lightweight disclaimers are operationally weak because they do not tell the user what kind of evidence boundary is binding, what claim family is affected, or what action should follow. A useful completeness layer must therefore be structured. It should help a consumer decide whether to trust, defer, escalate, request broader access, or switch to a human review path.

This also creates a concrete design pressure on agent architecture. Systems should preserve boundary metadata close to the answer surface rather than burying it in logs or access-control internals. Retrieval stacks, orchestration layers, and UI layers all need to cooperate if the final answer is going to remain interpretable as an answer under scope rather than an answer about the whole task.

\subsection{What Kinds of Agents This Benchmark Pressures}

The benchmark is especially relevant for four classes of systems.

\paragraph{Retrieval-backed assistants.}
These are the most direct case. They synthesize over authorized documents and can easily produce polished but incomplete outputs if missing evidence is not elevated into the answer protocol.

\paragraph{Delegated workflow agents.}
These systems act on behalf of a requester whose scope may differ from the system operator's scope or from adjacent sub-agents' scopes. For them, completeness errors can compound across handoffs, making the final answer appear more institutionally grounded than it actually is.

\paragraph{Incident and audit copilots.}
These settings are particularly sensitive because downstream humans often use the system output to decide whether to investigate further, disclose, block, or sign off. The cost of silent incompleteness is therefore not just an incorrect answer but a miscalibrated review path.

\paragraph{Multi-step planning agents.}
As systems become more tool-using and stateful, the benchmark's core issue becomes longitudinal. An agent may begin with a scoped evidence view, make an early incomplete inference, and then preserve that inference as if it were stable fact across later planning steps. That makes completeness awareness a state-management concern as well as an output concern.

\subsection{Deployment Tradeoffs Exposed by the Benchmark}

One reason the discussion section needs to be stronger than a normal benchmark paper is that the measured tradeoff is not merely ``accuracy versus abstention.'' The benchmark exposes at least three competing pressures:
\begin{enumerate}
    \item avoid unsafe polished overclaiming,
    \item avoid excessive conservative blocking, and
    \item preserve enough structured explanation for the output to remain operationally useful.
\end{enumerate}

Those pressures map onto real deployment choices. A security team may prefer cautious blocking in early triage, while a diligence workflow may prefer partial synthesis with explicit gap reporting so that human reviewers can continue efficiently. A single scalar metric is therefore not enough to support model or policy selection. The benchmark's multi-surface scoring is important precisely because different organizations will choose different points on that frontier.

This also implies that ``safer'' and ``better'' will not always move together automatically. A model that avoids false completeness by suppressing too many answerable cases may be unacceptable in high-throughput environments. A model that answers aggressively may look productive in demos while remaining institutionally unsafe. The benchmark's job is not to collapse those judgments into one universal score, but to make the tradeoffs legible enough that deployment teams can choose deliberately.

\subsection{Takeaways for Current Agentic AI}

For the current moment in agentic AI, four takeaways matter.
\begin{enumerate}
    \item \textbf{Retrieval quality is not the whole reliability story.} A system can retrieve and reason correctly over the documents it sees while still being unsafe about what it did not see.
    \item \textbf{Authorization and truth are different axes.} A compliant system is not automatically a trustworthy one if its outputs blur the line between authorized evidence and complete evidence.
    \item \textbf{Structured incompleteness is a capability, not a disclaimer.} Systems that can name the missing scope, affected claim, or blocked requirement are materially more useful than systems that only hedge in generic language.
    \item \textbf{Evaluation should track justified action, not only answer production.} Enterprise agent evaluation needs to move closer to the actual decision conditions under which systems are used. That means testing not only whether a system can answer, but whether it can preserve the relationship between scope, evidence, and justified action. Benchmarks that ignore that relationship risk producing reassuring scorecards for systems that are still brittle at the exact moment where organizational trust is won or lost.
\end{enumerate}

\section{Limitations}

This work has several limitations.

First, the corpora are synthetic. That is a feature for deterministic scoring, but it also means the benchmark compresses many difficulties of live enterprise evidence: messy formatting, long-range cross-document dependencies, latent authorship ambiguity, inconsistent naming, and uncertain document provenance. A strong result on this benchmark should therefore be interpreted as evidence of competence on a cleanly specified version of the problem, not as proof of robustness on live production corpora.

Second, the benchmark can in principle be gamed through template learning. The task families are intentionally structured, and a model or wrapper that memorizes their hidden-evidence patterns could improve benchmark performance without acquiring genuinely general completeness awareness. The current use of multiple templates and seeds helps, but it does not eliminate this risk. Future releases should broaden paraphrase variation, evidence topology, and scenario diversity to make template overfitting harder.

Third, the current real-model evidence is still partly transport- and provider-dependent. The cross-provider comparisons rely on mixed execution paths, prompt packaging, and serving layers. That is sufficient to show that the benchmark captures a real failure mode, but not sufficient to claim a final leaderboard or a stable ordering across model families. The provider-consistent Claude extension partially addresses this, but more transport-normalized comparisons are still needed.

Fourth, gap-report quality is only one operationalization of completeness awareness. We emphasize it because structured missing-scope reporting is operationally useful in enterprise settings, but other representations could also matter: calibrated completeness confidence, explicit escalation recommendations, workflow-specific refusal schemas, or UI-mediated human handoff patterns. A system that performs modestly on our gap-report metric may still embody a useful incompleteness protocol if it surfaces the right information in another form.

\section{Future Work}

Several extensions would make the benchmark materially stronger.

\paragraph{Broader scenario coverage.}
Future releases should add domains such as procurement, legal review, insider-risk investigation, and HR-sensitive workflows. The current three families show that the failure mode generalizes beyond one story, but they do not yet exhaust the space of consequential enterprise decisions.

\paragraph{Richer evidence topologies.}
The present benchmark uses compact six-document tasks because they are easy to score deterministically. A next step is to evaluate the same failure mode over longer chains of evidence, multi-hop requirements, contradictory subthreads, and temporally evolving corpora.

\paragraph{Interface-sensitive evaluation.}
The benchmark currently treats the model output as the main observation surface. In practice, the surrounding interface also matters. Future work should compare free-form answers, structured gap reports, refusal templates, escalation prompts, and human handoff mechanisms as distinct interventions over the same underlying tasks.

\paragraph{Training and control interventions.}
The benchmark is well-suited not only for evaluation but also for intervention studies. It can test whether prompt shaping, system-level policy scaffolding, explicit boundary metadata, or governance primitives such as fail-and-report reduce unsafe completeness without inducing excessive over-blocking.

\paragraph{Longitudinal agent testing.}
As agent systems become more stateful and delegated, evaluation should move beyond one-shot runs. A natural extension is to test whether agents preserve completeness awareness across multi-turn plans, delegated sub-requests, and changing authorization contexts.

\section{Conclusion}

Authorization-limited evidence is a distinct and measurable failure mode in agentic systems. A system can enforce permissions correctly and still mislead its consumer about result completeness. \texttt{partial-evidence-bench} turns that problem into a deterministic benchmark with explicit ACL partitions, completeness oracles, structured gap-report scoring, and scenario-sensitive evaluation.

The main practical result is simple: silent filtering is unsafe. The broader result is that enterprises need to reason not only about what a system may access, but also about how a system represents the limits of what it was allowed to see. As agentic AI moves deeper into delegated, scoped, document-grounded workflows, that distinction will matter more, not less. Benchmarks that surface completeness integrity under authorization constraints are therefore part of the core evaluation stack for serious enterprise agents, not an optional governance extra.

\end{document}